\documentclass[letterpaper, 10 pt, conference]{ieeeconf}  % Comment this line out if you need a4paper

\IEEEoverridecommandlockouts                              % This command is only needed if 
\usepackage{commath}
\usepackage{xcolor,colortbl}
\usepackage{graphicx}
\usepackage{subcaption}
\usepackage{adjustbox}

\usepackage{hyperref} % for url
\usepackage{textcomp, gensymb} % for degree symbol, trademark
\usepackage{siunitx}
\usepackage{amsmath}
\usepackage{tabularx}

\usepackage{tipa}

\usepackage{makecell} % multi line in a cell in a table

\title{\LARGE \bf
Single Actuator Undulation Soft-bodied Robots Using\\A Precompressed Variable Thickness Flexible Beam
}

\author{Tung D. Ta$^{1}$% <-this % stops a space
\thanks{*This work was supported by JSPS Grant-in-Aid for Scientific Research~(B) Grant Number 23H01376, Japan}% <-this % stops a space
\thanks{$^{1}$Graduate School of Information Science and Technology, The University of Tokyo, 7-3-1 Hongo, Bunkyo-ku, Tokyo, Japan \tt\small tung@csg.ci.i.u-tokyo.ac.jp}%
}

\begin{document}

\maketitle
\thispagestyle{empty}
\pagestyle{empty}

%%%%%%%%%%%%%%%%%%%%%%%%%%%%%%%%%%%%%%%%%%%%%%%%%%%%%%%%%%%%%%%%%%%%%%%%%%%%%%%%
\begin{abstract}
Soft robots - due to their intrinsic flexibility of the body - can adaptively navigate unstructured environments. One of the most popular locomotion gaits that has been implemented in soft robots is undulation. The undulation motion in soft robots resembles the locomotion gait of stringy creatures such as snakes, eels, and \textit{C.~Elegans}. Typically, the implementation of undulation locomotion on a soft robot requires many actuators to control each segment of the stringy body. The added weight of multiple actuators limits the navigating performance of soft-bodied robots. In this paper, we propose a simple tendon-driven flexible beam with only one actuator (a DC motor) that can generate a mechanical traveling wave along the beam to support the undulation locomotion of soft robots. The beam will be precompressed along its axis by shortening the length of the two tendons to form an S-shape, thus pretensioning the tendons. The motor will wind and unwind the tendons to deform the flexible beam and generate traveling waves along the body of the robot. We experiment with different pre-tension to characterize the relationship between tendon pre-tension forces and the DC-motor winding/unwinding. Our proposal enables a simple implementation of undulation motion to support the locomotion of soft-bodied robots.

\end{abstract}

%%%%%%%%%%%%%%%%%%%%%%%%%%%%%%%%%%%%%%%%%%%%%%%%%%%%%%%%%%%%%%%%%%%%%%%%%%%%%%%%
\section{INTRODUCTION}

% \textsf{\raisebox{.23ex}{\adjustbox{height=1.1ex}{(}}} % (
% \textsf{\raisebox{.23ex}{\adjustbox{height=1.1ex}{)}}} % )
% \textsf{\raisebox{.28ex}{\rotatebox[origin=c]{90}{$\sim$}}} % 2
% \textsf{\raisebox{.28ex}{\reflectbox{\rotatebox[origin=c]{90}{$\sim$}}}} % S
% \textsf{\raisebox{.04ex}{\adjustbox{height=1.26ex}{\textrevglotstop}}} % q
% \textsf{\raisebox{.04ex}{\adjustbox{height=1.26ex}{\reflectbox{\textrevglotstop}}}} % p
% \textsf{\rotatebox[origin=c]{180}{\raisebox{.04ex}{\adjustbox{height=1.26ex}{\textrevglotstop}}}} % b
% \textsf{\rotatebox[origin=c]{180}{\raisebox{.04ex}{\adjustbox{height=1.26ex}{\reflectbox{\textrevglotstop}}}}} % 

Soft robotics is paving the way for new thinking methodologies in the design, fabrication, and control of robots. Benefiting from their intrinsic flexibility, soft-bodied robots can adapt quickly to unstructured environments. They can deform their body, navigate confined spaces, and conduct tasks in search and rescue missions. 

One of the most popular ways for soft-bodied robots to move is to undulate. Inspired by the lateral undulation locomotion of snakes, many robots for navigation and exploration tasks take the form of a robotic snake that can undulate itself through the terrains~\cite{Ta2018-nu, Qin2018-dr, Wang2023-nn, Jia2023-qz, Qi2023-tc}. Locomotion by undulating can be seen in other forms such as anguilliform in sea eels~\cite{Tytell2004-ai, Leftwich2012-ki, Ta2020-ys}. One of the problems in making robots that can undulate through space is that the generation of undulation motion requires multiple actuators. For example, a rigid snake-like robot will need the same number of actuators as the number of joins it has. This added weight makes the robot heavier, complicating the design and control, and limiting the agility of the robots. The same story goes for soft-bodied snake-like robots. Though the form of actuators in soft robotics might be different (such as using pneumatic pumps instead of electric motors), soft snake robots still require controlling each body segment independently using separated actuators.

\begin{figure}[t]
    \centering
    \includegraphics[width = \linewidth]{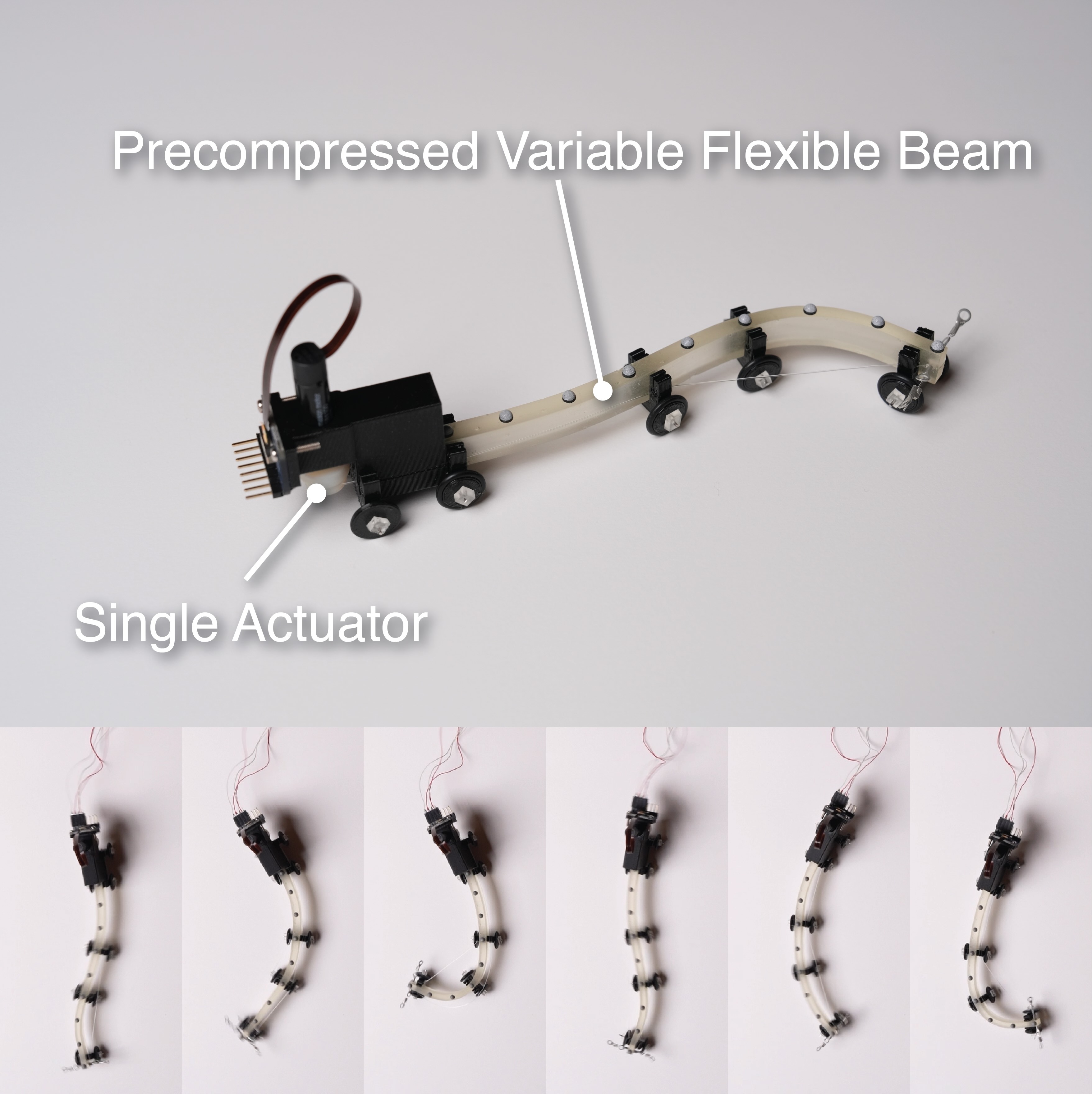} 
    \caption{Top: a 3D printed snake-like soft-bodied robot with a pre-compressed tendon-driven variable thickness flexible beam body. The robot uses only one actuator (a DC motor) to generate traveling mechanical waves along the body to support undulation locomotions. Bottom: snapshots of an undulation motion.}
    \label{fig:top}
\end{figure}

\subsection{Tendon-driven Soft Robots}
Tendon-driven robots are not only abanduntly popular in rigid robotics~\cite{Lens2013-xb, Rao2023-jy, In2015-ev} but also widely adopted in soft robotics~\cite{Zhang2022-zr, Meng2020-aq, Gunderman2022-kl}.

Synchronously pulling and releasing tendons enable many dexterous manipulations. The soft fingers in~\cite{Cheng2021-us} use the tendon to control a compliant mechanism to perform bending motions. The octopus-inspired soft robotic arm in~\cite{Laschi2012-ee} uses tendons to mimic the smooth and complicated movement of the octopus' tentacles. The soft wearable robot in~\cite{Kang2019-ny} proposes a special tendon routing mechanism to allow pinch and grasp motion for patients with hand disability.

In most of the cases, tendon-driven soft robots are actuated by motors. A motor can wind and unwind a tendon to pull and release it through a well-designed system of pulleys. Researchers are using tendons to make insect-inspired soft robots~\cite{Umedachi2019-qx} that can navigate the wild.

\subsection{Undulation in Soft-bodied Robots}
Undulation is one of the common ways for creatures such as snakes, lampreys, \textit{C. Elegans}, and sea eels to navigate the environments. Inspired by nature, many soft robots adopt the undulation motion for their locomotion. The soft robot in~\cite{Ta2020-ys} uses shape memory alloys to generate undulation of a stringy body to make the robot crawl on land and swim in viscous liquid environments. The soft robot in~\cite{Onal2013-lx} uses pneumatic actuators to mimic the locomotion of a snake. The biomimetic soft robot in~\cite{Xia2023-fm} mimics the fins of manta rays to move both underwater and on dry terrain.

Undulation motion can be generated by coordinating the actuation of a chain of actuators~\cite{Anastasiadis2023-sf} or by using one motor to actuate a specially designed mechanism. The single actuator wave-like robot in~\cite{Zarrouk2016-ml} uses a DC motor to convert rotation motion to undulation through a helical shaft.

In this paper, we propose a simple tendon-drive flexible beam that can generate undulation movement with only one DC motor. The key ideas are (a) to use a flexible beam with variable thickness and (b) to pre-compress the beam into an S-shaped. This configuration enables offloading of the undulation control onto the structure of the beam. Therefore, we need only one DC motor to control and generate a traveling mechanical wave along the body of the beam. This traveling mechanical wave, in turn, can be used to generate the undulation motion of slender beam-based soft robots. Our contributions include:
\begin{itemize}
  \item Designing of a tendon-driven variable thickness flexible beam that generates undulation motion using one actuator,
  \item Analysis and evaluation of the performance of the beam with different designing parameters such as variable thickness, precompression, and amplitude of the tendon's winding/unwinding.
  \item Demonstrate the undulation locomotion of the proposed flexible beam in a snake-like soft-bodied robot.
\end{itemize}

\section{DESIGN OF A TENDON-DRIVEN VARIABLE THICKNESS FLEXIBLE BEAM}
In this paper, we report the design and operation of a flexible tendon-driven beam that generates mechanical traveling waves along the axis of the beam to support the undulation locomotion of a snake-like soft-bodied robot. Design decisions include body shape, initial precompression magnetitude, and actuation mechanism.

\begin{figure}[t]
    \centering
    \includegraphics[width = \linewidth]{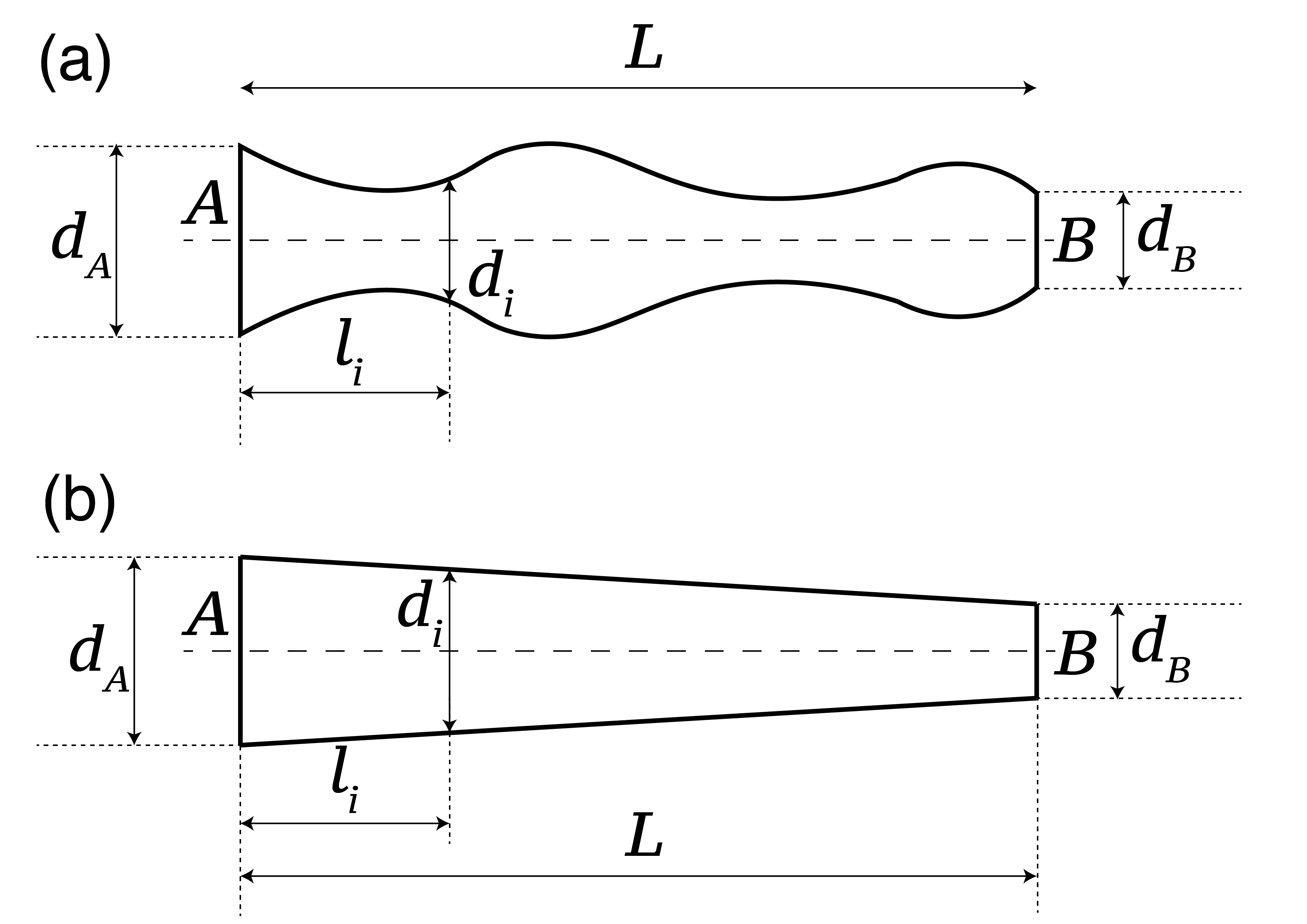}
    \caption{Parametric design of the variable thickness flexible beam, (a) nonlinear variable thickness, (b) linear variable thickness.}
    \label{fig:beam2d-design}
\end{figure}
\subsection{Design of Variable Thickness Flexible Beams}\label{sec:beam-design}
A variable-thickness flexible beam has different thicknesses at different points along the axis of the beam. We will limit our scope to beams that are symmetric through its neutral axial. Taking into account the length of the beam $L$, let $AB$ be the neutral axis of the beam (as shown in Fig.~\ref{fig:beam2d-design}). The thicknesses of the beam at $A$ and $B$ are $d_A \geq 0$ and $d_B \geq 0$. The thickness of the beam at a point $i$ in the segment $AB$, with the distance $l_i$ from A, is the following:

\begin{equation}
d_i = f(l_i)\text{ with }
    \begin{cases}
      0 \leq \l_i \leq L\\
      d_i > 0
    \end{cases}
\end{equation}

The function $f$ can be adjusted to form different morphologies of the variable thickness flexible beams. In this paper, we choose a simple linear function (as shown in Fig.~\ref{fig:beam2d-design}b) to make a tapered flexible beam as follows:

\begin{equation}
d_i = f(l_i) = \frac{d_A - d_B}{L}l_i + d_B \text{ with }
    \begin{cases}
      0 \leq \l_i \leq L\\
      d_i > 0
    \end{cases}
\end{equation}
The values of morphology parameters are shown in TABLE~\ref{tab:params_design}
\begin{table}[b]
    \caption{Parametric design of variable thickness flexible beams}
    \label{tab:params_design}
    \begin{center}
        \begin{tabular}{r|l}
            \textbf{Parameter} & \textbf{Value}\\
            \hline \hline
            Beam length $L$ & $L=\SI{140}{\milli\metre}$  \\
            \hline
            Beam height $H$ & $H=\SI{10}{\milli\metre}$  \\
            \hline
            Thickness $d_A$ & $d_A=\SI{6}{\milli\metre}$\\
            \hline
            Thickness $d_B$ & $d_B=\{$\SI{6}{\milli\metre}$, $\SI{4}{\milli\metre}$, $\SI{2}{\milli\metre}$\}$\\
            \hline
            Tendon length offset $\Delta_L$ & $\Delta_L = \{$\SI{0}{\milli\metre}$, $\SI{5}{\milli\metre}$, $\SI{10}{\milli\metre}$, $\SI{15}{\milli\metre}$\}$\\
            \hline
            Tendon wind/unwind $\Delta_\tau$ & $\Delta_\tau = \makecell{\{$\SI{15}{\milli\metre}$, $\SI{20}{\milli\metre}$, $\SI{25}{\milli\metre}$,\\$\SI{30}{\milli\metre}$, $\SI{35}{\milli\metre}$, $\SI{40}{\milli\metre}$\}}$\\
            \hline
        \end{tabular}
    \end{center}
\end{table}

\subsection{Pre-compressed Flexible Beams}
While actuating a flexible beam, the undulation motions will emerge if there is a traveling mechanical wave along the body of the beam. In other words, the phase-shift $\theta$ of the body displacement between different points on the beam needs to be $0< \theta < 180$. This can be achieved by using two motors with nonzero phase shift to control each half-length segment of the beam~\cite{Ta2018-nu}.

\begin{figure}[t]
    \centering
    \includegraphics[width = \linewidth]{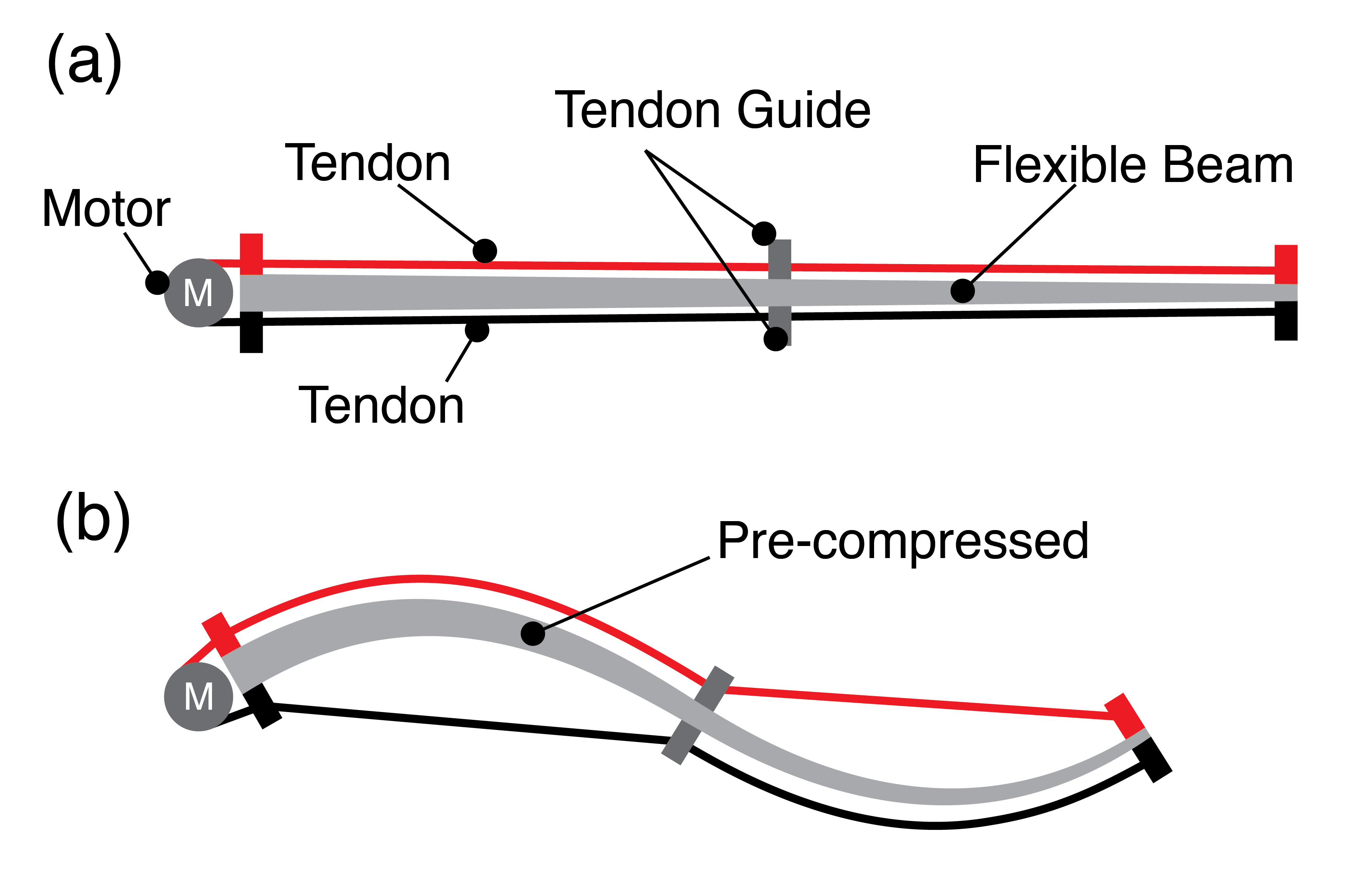}
    \caption{Design of the variable thickness flexible beam. One end of the beam is tapered. The thicknesses of the two ends of the beam are \SI{6}{\milli\metre}-\SI{6}{\milli\metre}, \SI{6}{\milli\metre}-\SI{4}{\milli\metre}, \SI{6}{\milli\metre}-\SI{2}{\milli\metre}. The beam is initially pre-compressed by connecting two tendons to a motor at the thickest end. (a) Tapered flexible beam with tendons, (b) Pre-compressed flexible beam by shortening both tendons to induce initial buckling state.}
    \label{fig:initial-buckle-design}
\end{figure}

We aim to design a flexible beam that can generate undulation motions with just one motor as an actuator. This goal poses a challenge of how to offload the phase-shift control from motor control to the morphology structure of the beam. Besides using variable thickness flexible beams, we notice that precompressing this beam to buckle the beam into an S-shaped could help incorporate the phase shift of the traveling mechanical wave into the morphology structure of the beam. We adjust the precompression rate by changing the initial tendons' length with an offset $\Delta_L$ from the beam length $L$. The initial length of the two tendons is $L_{T} = L - \Delta_L$. The initial buckling S-shaped beam is shown in Fig.~\ref{fig:initial-buckle-design}.

To prevent tendons from crossing the neutral axis during actuation, we add a tendon guide to the middle point of the flexible beam (as shown in Fig.~\ref{fig:initial-buckle-design}).

\begin{figure}[t]
      \centering
      \includegraphics[width=\linewidth]{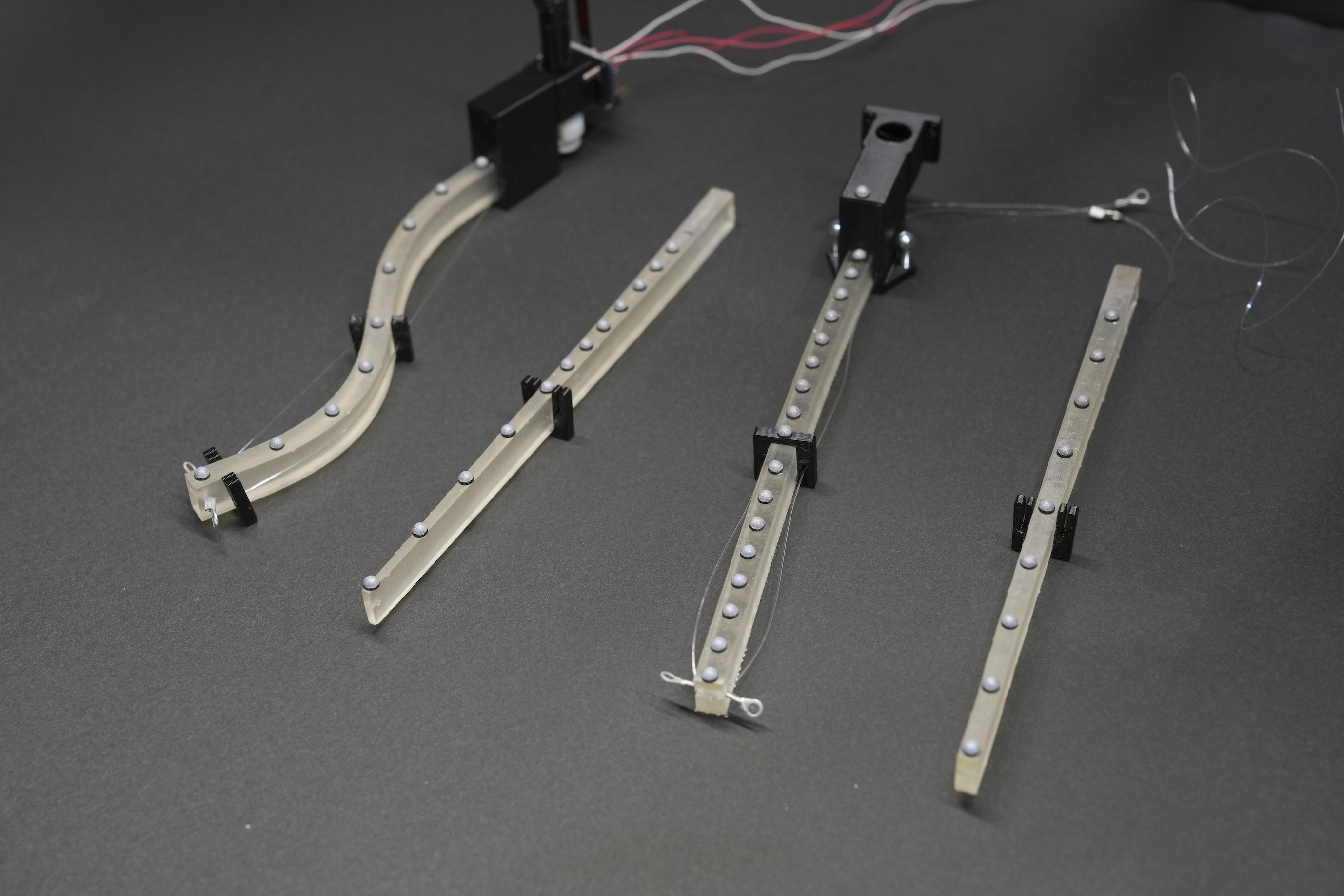}
      \caption{3D-printed variable thickness flexible beam with different value of morphology parameters}
      \label{fig:3dprinted}
\end{figure}
\begin{table}[t]
    \caption{Fabrication Information}
    \label{tab:fab_info}
    \begin{center}
        \begin{tabular}{r|l}
            \textbf{Device} & \textbf{Name}\\
            \hline \hline
            SLA 3D printer & Form~3+  \\
            \hline
            Flexible Material & Elastic~50A\\
            \hline
            FDM 3D printer & Prusa~i3~MK3S+  \\
            \hline
            Rigid Material & PETG\\
            \hline
            Motor & Maxon 347726 DC Motor \SI{8}{\milli\metre} \SI{0.5}{\watt} \\
            \hline
            Tendon & Nylon fishing wire $\phi = \SI{0.4}{\milli\metre}$\\
            \hline
        \end{tabular}
    \end{center}
\end{table}

\begin{figure}[t]
      \centering
      \includegraphics[width=\linewidth]{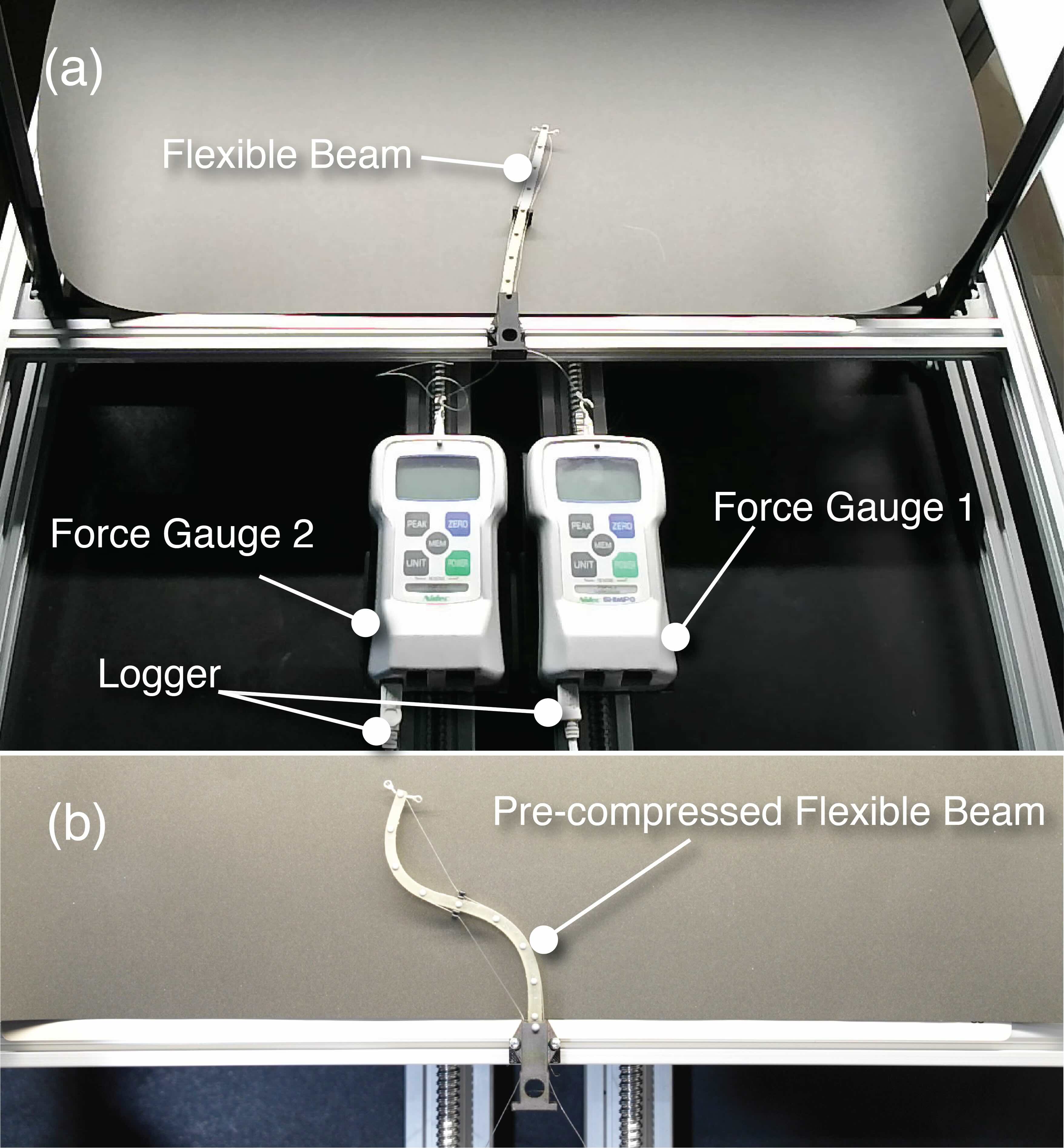}
      \caption{(a) We measure the tension force of the tendons using two hand-held force gauges. We simulate the winding/unwinding of the motor by moving the two hand-held force gauges in reversed order. We experiment one by one with three types of variable thickness flexible beam $S_{62}$, $S_{64}$, $S_{66}$. (b) Each test will start by precompressing the flexible beam.}
      \label{fig:force_test_setup}
\end{figure}

\subsection{Actuation by One Motor}
As mentioned above, we will use only one motor as an actuator of the flexible beam. The motor is attached to one end of the beam with a double pulley that winds and unwinds the tendons as illustrated in Fig.~\ref{fig:initial-buckle-design}. As the phase-shift controlling mechanism has been offloaded to the morphology structure of the beam, the motor just needs to wind/unwind the tendons by alternatively rotating clockwise and anti-clockwise with a fixed frequency. The motor can control the shape of the undulation motion by changing the amount of tendon winding/unwinding $\Delta\tau$ (as shown in TABLE~\ref{tab:params_design}).
\begin{figure*}[t]
      \centering
      \includegraphics[width=.95\linewidth]{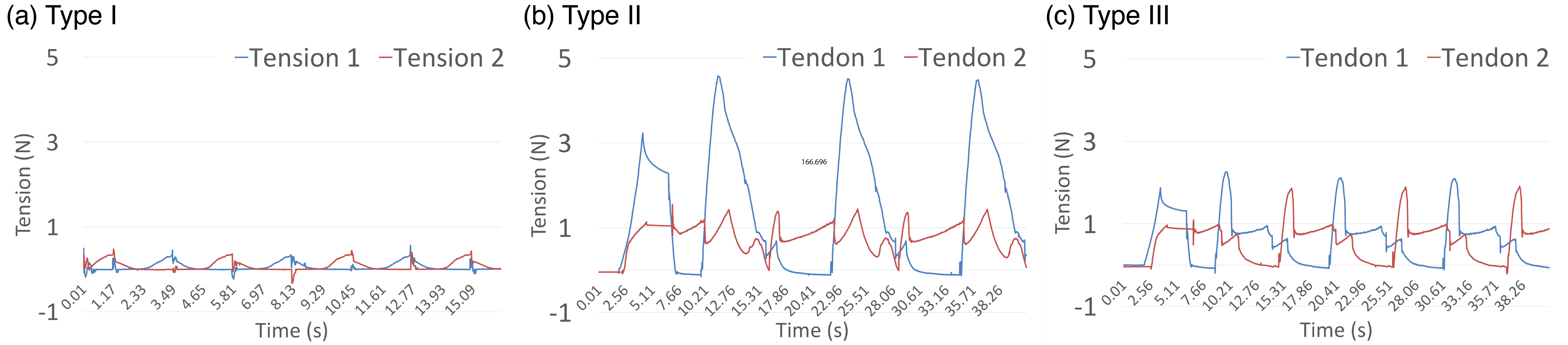}
      \caption{The result of measuring tension on each tendon during actuation (the first 4 times in total of 10 times repetition). (a) TYPE I: Simple bending (sample $S_{64}, \Delta_L = 0, \Delta_\tau = 15$), (b) TYPE II: Simple bending with buckling (sample $S_{64}, \Delta_L = 15, \Delta_\tau = 35$), (c) TYPE III: Buckling with traveling mechanical wave (sample $S_{64}, \Delta_L = 10, \Delta_\tau = 35$)}
      \label{fig:tension_result}
\end{figure*}

\subsection{Fabrication of the Variable Thickness Flexible Beams Sample}
Our flexible beams are made of deformable materials such as rubber and silicone. For fabrication, we use a Stereolithography (SLA) 3D printer (Form~3+) to print the flexible beams with an elastic material (Elastic~50A) from Formlabs\footnote{\url{https://formlabs.com/}}. The rigid parts such as tendon guides, pulleys, and motor housing are printed using a commonly available Fused Deposition Modeling (FDM) 3D printer (Prusa~i3~MK3S+) from Prusa Research\footnote{\url{https://www.prusa3d.com/}} with Polyethylene terephthalate glycol (PETG) filaments from Overture\footnote{\url{https://overture3d.com/}}. We use a Maxon\footnote{\url{https://www.maxongroup.com/}} DC motor (Maxon 347726 DC Motor \SI{8}{\milli\metre} \SI{0.5}{\watt}) to actuatate the flexible beams. We use nylon fishing wires\footnote{\url{https://tegusuya.jp/}} with diameter $\phi = \SI{0.4}{\milli\metre}$ to make two tendons that connect the beam and the motor. The details of the fabrication devices and materials are listed in TABLE~\ref{tab:fab_info}. Fig.~\ref{fig:3dprinted} shows samples of the 3D-printed variable thickness flexible beams.

\section{EXPERIMENT}\label{sec:experiment}
This section will evaluate the performance of the pre-compressed variable thickness flexible beam in generating undulation motion using only one motor. We will measure the tension force on each tendon during the winding/unwinding process. We will also study the displacement of multiple points along the beam, as well as the body shape of the beam while being actuated. In all experiments, we have three types of variable-thickness flexible beams, as follows:
\begin{itemize}
    \item $S_{62}$: $d_A = \SI{6}{\milli\metre}$, $d_B = \SI{2}{\milli\metre}$
    \item $S_{64}$: $d_A = \SI{6}{\milli\metre}$, $d_B = \SI{4}{\milli\metre}$
    \item $S_{66}$: $d_A = \SI{6}{\milli\metre}$, $d_B = \SI{6}{\milli\metre}$
\end{itemize}

\subsection{Tendon Tension Experiment}\label{sec:tendon_tension_exp}
This experiment aims to measure the tension force on each tendon while the flexible beam is actuated.

\subsubsection{Experiment Setup}
As shown in Fig.~\ref{fig:force_test_setup}, we use two hand-held force gauges Nidec FGP-0.5\footnote{\url{https://www.nidec.com/}} to measure the tension force of the tendons during actuation. The two force gauges are screwed to two linear rails driven by two NEMA~17 step motors. The two tendons are then attached to the force gauges using two rounded hooks. The force gauges will log the measurement to a computer with a measurement frequency of \SI{100}{\hertz}.

In this experiment, we approximately mimic the winding/unwinding motion of a motor by moving the two force gauges in reversed order with the same displacement distance. Such motion will alternatively pull and release the two tendons at the same time with the same amount of displacement.

For each sample $S_{62}$, $S_{64}$, and $S_{66}$, we experiment with different values of initial tendon offset $\Delta_L$ and winding/unwinding displacement $\Delta_\tau$, as follows:
\begin{itemize}
    \item $\Delta_L = \{\SI{0}{\milli\metre}, \SI{5}{\milli\metre}, \SI{10}{\milli\metre}, \SI{15}{\milli\metre}\}$
    \item $\Delta_\tau = \{\SI{15}{\milli\metre}, \SI{20}{\milli\metre}, \SI{25}{\milli\metre}, \SI{30}{\milli\metre}, \SI{35}{\milli\metre}, \SI{40}{\milli\metre}\}$
\end{itemize}

For each experiment, we repeat the measurement $N$ times wiht $N = 10$. The procedure of each measurement includes:
\begin{itemize}
    \item Place the flexible beam in resting state without any tension or compression
    \item Start logging the tension force readings
    \item Shorten the two tendons with the same amount of tendon initial offset $\Delta_L$ by moving the two force gauges backward a distance $\Delta_L$
    \item Start moving the two force gauges in inverse direction with the distance of $\Delta_\tau$ to simulate the winding/unwinding of a motor, repeat this step for $N = 10$ times
\end{itemize}

\subsubsection{Experiment Result}
\begin{figure*}[t]
      \centering
      \includegraphics[width=\linewidth]{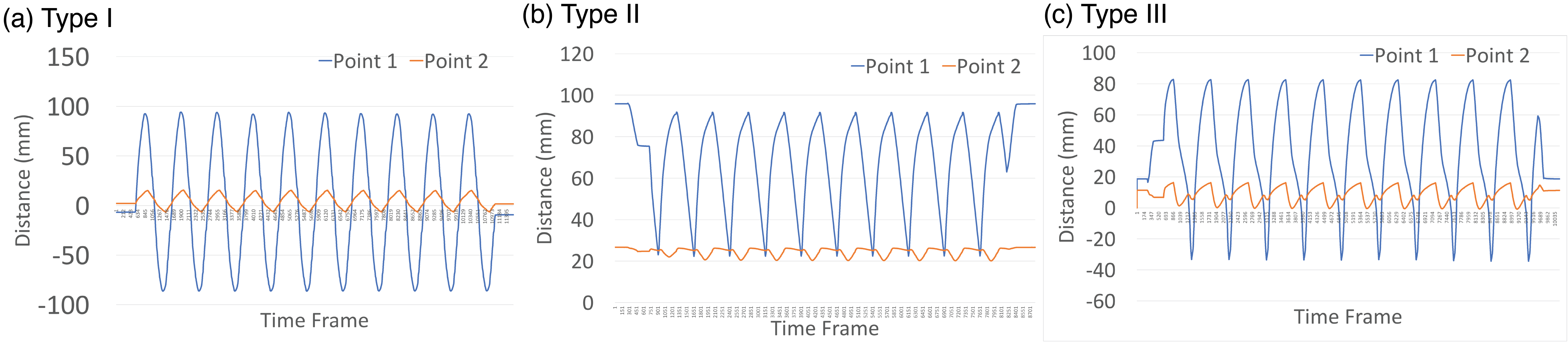}
      \caption{Displacement of multiple points along the body of the flexible beam}
      \label{fig:mocap}
\end{figure*}
\begin{figure}[t]
      \centering
      \includegraphics[width=\linewidth]{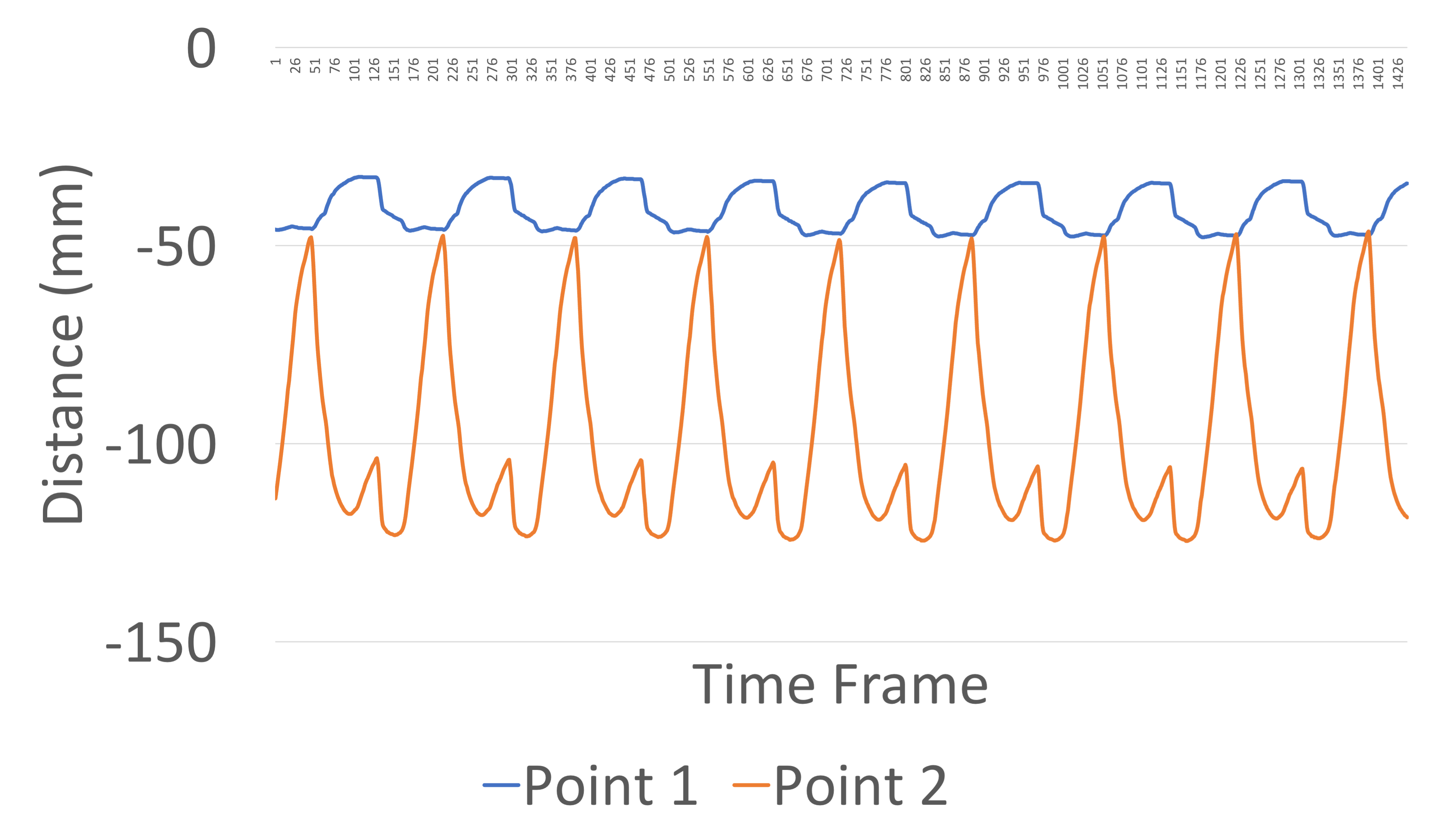}
      \caption{Motion capture of a precompressed variable thickness flexible beam actuated by a motor.}
      \label{fig:mocap_motor}
\end{figure}
A part of the result of the tension force test is shown in Fig.~\ref{fig:tension_result}. From the data, we notice that there are three patterns of the motion of the beam with three different tension profiles.

\underline{TYPE I: Simple bending:} This happens when $\Delta_L = 0$. In this case, the initial tendon offset $\Delta_L = \SI{0}{\milli\metre}$, the tension on both tendons are \SI{180}{\degree} out of phase to each other. This signals that the beam simply bends from one side to the other side without generating any traveling mechanical wave along the flexible beam. The graph in Fig.~\ref{fig:tension_result} for the case of $\Delta_L = 0$ does not show any sudden drop of tension on both tendons. The peak of tensions of both tendons are also relatively low, comprared to the other two types. This implies that there is no buckling happening in this case.

\underline{TYPE II: Simple bending with buckling:} This happens when $\Delta_L > 0$ and $\Delta_\tau$ is small. In this case, the initial tendon offset $\Delta_L > 0$, but the winding/unwinding $\Delta_\tau$ of the tendons is still too small to produce a traveling mechanical wave. However, because $\Delta_L > 0$, there is compression inside the flexible beam that causes buckling. The peak of tensions of one tendon is relatively high compared to the peak of tension of the another tendon.

\underline{TYPE III: Buckling with traveling mechanical wave:} This happens when $\Delta_L > 0$ and $\Delta_\tau$ is large enough. The behavior of the flexible beam starts changing when the initial tendon offset $\Delta_L > 0$. In the case of $\Delta_L = \SI{10}{\milli\metre}$ and $\Delta_\tau = \SI{35}{\milli\metre}$, we can see that the tensions on the two tendons are out of phase with the phase change $\theta < \SI{180}{\degree}$. Additionally, there are multiple drops of tension on both tendons. The peak of tensions of both tendons are roughly the same. These clues signal that there is a traveling mechanical wave along the axis of the beam and there are local snap-through bucklings in the flexible beam. When comparing the tension data and the video data, we notice that the tension drops on the two tendons correspond to the snap-through bucklings of the half-beam from point $A$ to tension guide and from tension guide to point $B$.

\subsection{Body Shape Experiment}
In this section, we evaluate the body shape of the precompressed variable thickness flexible beams while actuated.

\subsubsection{Experiment Setup}

This experiment is carried out at the same time as the tension force measurement. We use a 3-camera (Flex~13 at \SI{120}{\hertz}) motion capture system from Optitrack\footnote{\url{https://optitrack.com/}} to capture the motion of multiple \SI{3}{\milli\metre}-semisphere-markers that are attached to the body of the flexible beam.

\subsubsection{Experiment Result}

The motion capture result is shown in Fig.~\ref{fig:mocap}. We show the data of the three types mentioned in the previous experiment.

The motion capture data of TYPE~I motion shows that the points on the body of the beam displace with different amplitudes but are in phase with each other. That means there is no traveling mechanical wave in this case.

The points on the body of the beam in the case of TYPE~II motion are slightly out-of-phase to each other. We also notice a slight bump in the position of some points when those points undergo a snap-through buckling. There is no efficient traveling mechanical wave in this TYPE~II either.

The points on the body of the beam in the case of TYPE~III motion are out-of-phase to each other. This clue shows that there are efficient traveling mechanical waves along the body of the beam. This traveling mechanical wave is strong enough to support the undulation locomotion of a snake-like soft-bodied robot that will be demonstrated at the end of this article. In the motion-capture data, we also notice slight bumps in the position of some points when those points undergo a snap-through buckling. The out-of-phase of TYPE~III motion is even more vivid when we actuate the flexible beam with a motor (as shown in Fig.~\ref{fig:mocap_motor}).

The results of the experiment in both tension force and motion capture show that by a simple change in the thickness of the flexible beam with a well-tuned precompression, a single motor can easily generate traveling mechanical waves to support undulation locomotion. We demonstrate this in the following part, reporting a snake-like soft-bodied robot with a motor as its single actuator.

\subsection{Demonstration of Single Actuator Snake-like Soft-bodied Robot}
\begin{figure}[t]
      \centering
      \includegraphics[width=\linewidth]{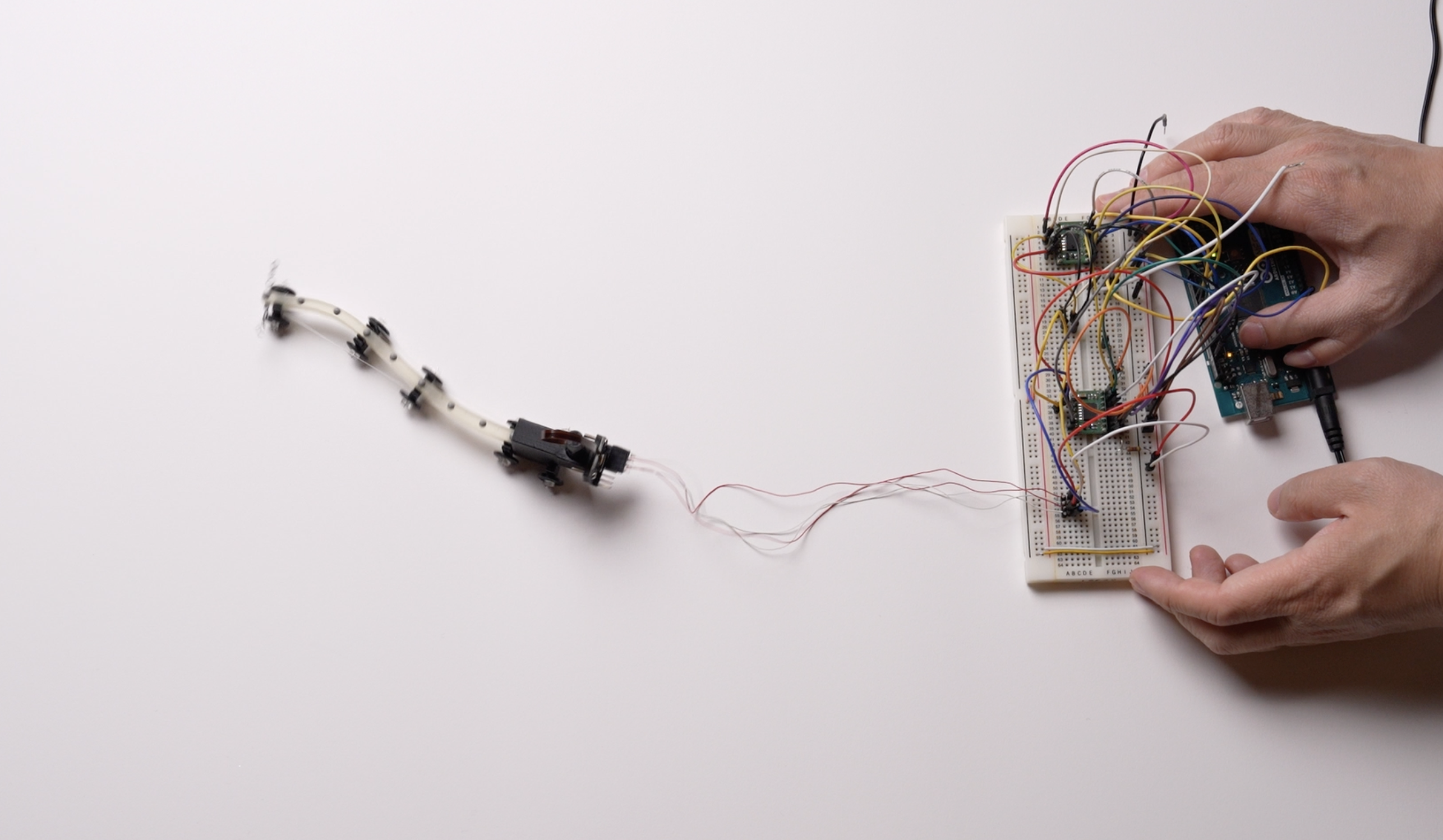}
      \caption{A single actuator snake-like soft-bodied robot running by generating undulation motion.}
      \label{fig:robot_run}
\end{figure}

\begin{figure}[t]
      \centering
      \includegraphics[width=\linewidth]{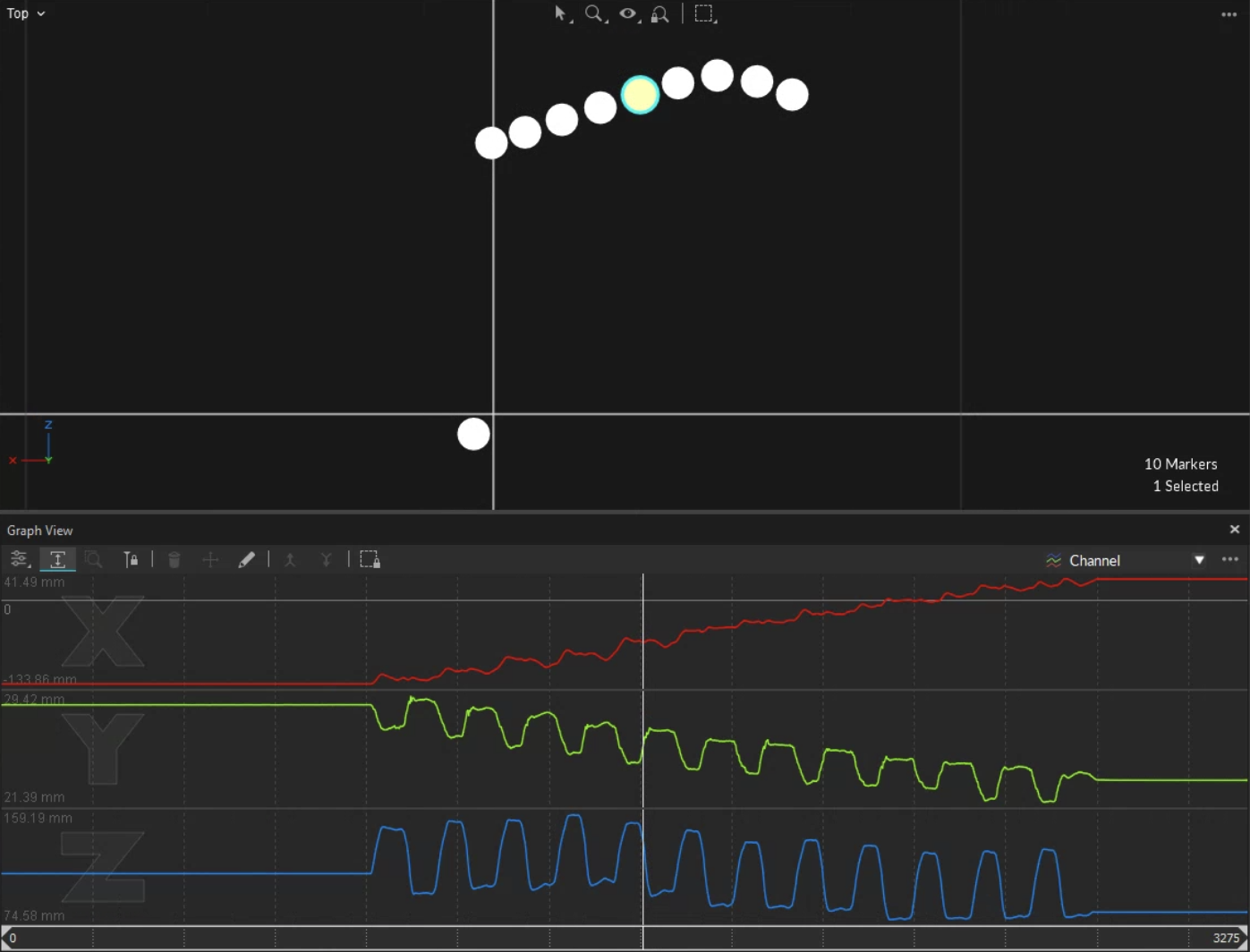}
      \caption{Motion capture of a single actuator snake-like soft-bodied robot undulating on a flat surface.}
      \label{fig:robot_run_mocap}
\end{figure}

Based on what we learned from the experiment above, we developed a simple two-segment snake-like soft-bodied robot. The body of the robot is the same as the experiment sample $S_{64}$. The undulation locomotion of a snake robot is heavily dependent on the anisotropic frictional surface between the ventral side of the robot and the ground. It is possible to print an anisotropic frictional robotic skin as shown in~\cite{Ta2018-nu, Ta2023-jp}. However, in the implementation this time, we will simply use passive wheels to realize the anisotropic frictional effect between the robot and the ground. Fig.~\ref{fig:robot_run} shows the single actuator snake-like soft-bodied robot undulating on a flat surface. Based on the motion capture data shown in Fig.~\ref{fig:robot_run_mocap}, it is clear that the robot moves effectively forward by undulating its entire body using a single actuator.


\section{DISCUSSION AND FUTURE WORKS}
By simply changing the thickness of a flexible beam along its axis and applying precompression on the beam, we can generate traveling mechanical waves along the body of the beam, thus, supporting undulation motions. The offloading of undulation control to the morphology structure of the flexible beam drastically simplifies the design of the actuator and control mechanism. Such an undulation is useful not only for locomotion on land but also for aquatic locomotions such as anguilliform maneuvering.

In the future, we are aiming to improve the performance and controllability of the pre-compressed variable thickness flexible beam for soft robotics, as follows:
\begin{itemize}
    \item \emph{Variable stiffness:} Changing the thickness of the beam effectively changes the local stiffness of the beam along its axis. In the future, we will incorporate variable stiffness with variable thickness to expand the designing space of the precompressed flexible beams in soft robotic design.
    \item \emph{Non-linear thickness function:} In this paper, we simplified our decision-making process by using a linear function for the variable thickness of the beam. In the future, we can explore other nonlinear functions to design the thickness/stiffness of the flexible beam to see how the change in shape affects the performance of the undulation motions.
    \item \emph{Application of undulation in different environments :} We demonstrated the application of single actuator pre-compressed variable thickness flexible beams by making a simple snake-like soft-bodied robot that can crawl on flat and dry surfaces. We can expand this idea to other forms of locomotion, such as the anguilliform form in~\cite{Ta2020-ys}. We can also take advantage of snap-through buckling in the pre-compressed flexible beam to develop soft robots that can perform agile maneuvers.
    % \item \emph{Untethering:} 
    
\end{itemize}
\section{CONCLUSIONS}
In this paper, we proposed an approach to design single actuator tendon-driven undulation soft-bodied robots. The ideas include designing a variable-thickness flexible beam and precompressing this beam to take advantage of the intrinsic elasticity in the flexible beam for generating undulation motions. By offloading the complexity of undulation generation to the morphology structure of the beam, we drastically simplified the design and control of actuators in undulation soft-bodied robots. We evaluated the performance of the beam in generating traveling mechanical waves depending on the variable thickness, precompression rate, and amplitude of tendon winding/unwinding. We also demonstrated the application of our pre-compressed variable thickness flexible beam in making a single-actuator snake-like soft-bodied robot that can undulate on flat surfaces. We envision using this structure in other forms of movement in different spaces such as wet and viscous environments.

\addtolength{\textheight}{-13.5cm}   % This command serves to balance the column lengths

%%%%%%%%%%%%%%%%%%%%%%%%%%%%%%%%%%%%%%%%%%%%%%%%%%%%%%%%%%%%%%%%%%%%%%%%%%%%%%%%

%%%%%%%%%%%%%%%%%%%%%%%%%%%%%%%%%%%%%%%%%%%%%%%%%%%%%%%%%%%%%%%%%%%%%%%%%%%%%%%%

%%%%%%%%%%%%%%%%%%%%%%%%%%%%%%%%%%%%%%%%%%%%%%%%%%%%%%%%%%%%%%%%%%%%%%%%%%%%%%%%

\section*{ACKNOWLEDGMENT}
This work was supported by JSPS Grant-in-Aid for Scientific Research~(B) Grant Number 23H01376, Japan. We thank Nobuyuki Umetani and Tetsuro Yamazaki for helping with the video shooting.

\bibliographystyle{IEEEtran}
% \bibliography{IEEE, paperpile} 
\bibliography{paperpile}

\begin{thebibliography}{10}
\providecommand{\url}[1]{#1}
\csname url@samestyle\endcsname
\providecommand{\newblock}{\relax}
\providecommand{\bibinfo}[2]{#2}
\providecommand{\BIBentrySTDinterwordspacing}{\spaceskip=0pt\relax}
\providecommand{\BIBentryALTinterwordstretchfactor}{4}
\providecommand{\BIBentryALTinterwordspacing}{\spaceskip=\fontdimen2\font plus
\BIBentryALTinterwordstretchfactor\fontdimen3\font minus \fontdimen4\font\relax}
\providecommand{\BIBforeignlanguage}[2]{{%
\expandafter\ifx\csname l@#1\endcsname\relax
\typeout{** WARNING: IEEEtran.bst: No hyphenation pattern has been}%
\typeout{** loaded for the language `#1'. Using the pattern for}%
\typeout{** the default language instead.}%
\else
\language=\csname l@#1\endcsname
\fi
#2}}
\providecommand{\BIBdecl}{\relax}
\BIBdecl

\bibitem{Ta2018-nu}
T.~D. Ta, T.~Umedachi, and Y.~Kawahara, ``Design of frictional {2D}-anisotropy surface for wriggle locomotion of printable soft-bodied robots,'' in \emph{2018 IEEE International Conference on Robotics and Automation (ICRA)}.\hskip 1em plus 0.5em minus 0.4em\relax IEEE, May 2018, pp. 6779--6785.

\bibitem{Qin2018-dr}
Y.~Qin, Z.~Wan, Y.~Sun, E.~H. Skorina, M.~Luo, and C.~D. Onal, ``Design, fabrication and experimental analysis of a 3-{D} soft robotic snake,'' in \emph{2018 IEEE International Conference on Soft Robotics (RoboSoft)}.\hskip 1em plus 0.5em minus 0.4em\relax IEEE, Apr. 2018, pp. 77--82.

\bibitem{Wang2023-nn}
T.~Wang, C.~Pierce, V.~Kojouharov, B.~Chong, K.~Diaz, H.~Lu, and D.~I. Goldman, ``\BIBforeignlanguage{en}{Mechanical intelligence simplifies control in terrestrial limbless locomotion},'' \emph{\BIBforeignlanguage{en}{Sci Robot}}, vol.~8, no.~85, p. eadi2243, Dec. 2023.

\bibitem{Jia2023-qz}
W.~Jia, Z.~Zhao, W.~Huang, Y.~Li, J.~Ling, B.~Chen, and Y.~Shen, ``Snake-inspired swarm robot design for distributed underwater search and rescue,'' in \emph{2023 IEEE International Conference on Robotics and Biomimetics (ROBIO)}.\hskip 1em plus 0.5em minus 0.4em\relax IEEE, Dec. 2023, pp. 1--6.

\bibitem{Qi2023-tc}
X.~Qi, T.~Gao, and X.~Tan, ``\BIBforeignlanguage{en}{Bioinspired {3D}-printed snakeskins enable effective serpentine locomotion of a soft robotic snake},'' \emph{\BIBforeignlanguage{en}{Soft Robot}}, vol.~10, no.~3, pp. 568--579, Jun. 2023.

\bibitem{Tytell2004-ai}
E.~D. Tytell and G.~V. Lauder, ``\BIBforeignlanguage{en}{The hydrodynamics of eel swimming: {I}. wake structure},'' \emph{\BIBforeignlanguage{en}{J. Exp. Biol.}}, vol. 207, no. Pt 11, pp. 1825--1841, May 2004.

\bibitem{Leftwich2012-ki}
M.~C. Leftwich, E.~D. Tytell, A.~H. Cohen, and A.~J. Smits, ``\BIBforeignlanguage{en}{Wake structures behind a swimming robotic lamprey with a passively flexible tail},'' \emph{\BIBforeignlanguage{en}{J. Exp. Biol.}}, vol. 215, no. Pt 3, pp. 416--425, Feb. 2012.

\bibitem{Ta2020-ys}
T.~D. Ta, T.~Umedachi, and Y.~Kawahara, ``A multigait stringy robot with bi-stable soft-bodied structures in multiple viscous environments,'' in \emph{2020 IEEE International Conference on Intelligent Robots and Systems (IEEE IROS’20)}.\hskip 1em plus 0.5em minus 0.4em\relax IEEE, Oct. 2020, pp. 8765--8772.

\bibitem{Lens2013-xb}
T.~Lens and O.~von Stryk, ``Design and dynamics model of a lightweight series elastic tendon-driven robot arm,'' in \emph{2013 IEEE International Conference on Robotics and Automation}, May 2013, pp. 4512--4518.

\bibitem{Rao2023-jy}
P.~Rao, C.~Pogue, Q.~Peyron, E.~Diller, and J.~Burgner-Kahrs, ``Modeling and analysis of tendon-driven continuum robots for rod-based locking,'' \emph{IEEE Robot. Autom. Lett.}, vol.~8, no.~6, pp. 3126--3133, Jun. 2023.

\bibitem{In2015-ev}
H.~In, B.~B. Kang, M.~Sin, and K.-J. Cho, ``Exo-glove: A wearable robot for the hand with a soft tendon routing system,'' \emph{IEEE Robot. Autom. Mag.}, vol.~22, no.~1, pp. 97--105, Mar. 2015.

\bibitem{Zhang2022-zr}
Y.~Zhang, W.~Zhang, J.~Yang, and W.~Pu, ``\BIBforeignlanguage{en}{Bioinspired soft robotic fingers with sequential motion based on tendon-driven mechanisms},'' \emph{\BIBforeignlanguage{en}{Soft Robot}}, vol.~9, no.~3, pp. 531--541, Jun. 2022.

\bibitem{Meng2020-aq}
J.~Meng, L.~Gerez, J.~Chapman, and M.~Liarokapis, ``A tendon-driven, preloaded, pneumatically actuated, soft robotic gripper with a telescopic palm,'' in \emph{2020 3rd IEEE International Conference on Soft Robotics (RoboSoft)}, May 2020, pp. 476--481.

\bibitem{Gunderman2022-kl}
A.~L. Gunderman, J.~A. Collins, A.~L. Myers, R.~T. Threlfall, and Y.~Chen, ``Tendon-driven soft robotic gripper for blackberry harvesting,'' \emph{IEEE Robotics and Automation Letters}, vol.~7, no.~2, pp. 2652--2659, Apr. 2022.

\bibitem{Cheng2021-us}
C.~Cheng, Y.~Yan, M.~Guan, J.~Zhang, and Y.~Wang, ``Tactile sensing with a tendon-driven soft robotic finger,'' in \emph{2021 9th International Conference on Control, Mechatronics and Automation (ICCMA)}.\hskip 1em plus 0.5em minus 0.4em\relax IEEE, Nov. 2021, pp. 14--19.

\bibitem{Laschi2012-ee}
C.~Laschi, M.~Cianchetti, B.~Mazzolai, L.~Margheri, M.~Follador, and P.~Dario, ``Soft robot arm inspired by the octopus,'' \emph{Adv. Robot.}, vol.~26, no.~7, pp. 709--727, Jan. 2012.

\bibitem{Kang2019-ny}
B.~B. Kang, H.~Choi, H.~Lee, and K.-J. Cho, ``Exo-glove poly {II}: A polymer-based soft wearable robot for the hand with a tendon-driven actuation system,'' \emph{Soft Robotics}, vol.~6, no.~2, pp. 214--227, Apr. 2019.

\bibitem{Umedachi2019-qx}
T.~Umedachi, M.~Shimizu, and Y.~Kawahara, ``Caterpillar-inspired crawling robot using both compression and bending deformations,'' \emph{IEEE Robotics and Automation Letters}, vol.~4, no.~2, pp. 670--676, Apr. 2019.

\bibitem{Onal2013-lx}
C.~D. Onal and D.~Rus, ``Autonomous undulatory serpentine locomotion utilizing body dynamics of a fluidic soft robot,'' \emph{Bioinspiration \& Biomimetics}, vol.~8, no.~2, pp. 026\,003--026\,010, Jun. 2013.

\bibitem{Xia2023-fm}
M.~Xia, H.~Wang, Q.~Yin, J.~Shang, Z.~Luo, and Q.~Zhu, ``Design and mechanics of a composite wave-driven soft robotic fin for biomimetic amphibious robot,'' \emph{J. Bionic Eng.}, vol.~20, no.~3, pp. 934--952, May 2023.

\bibitem{Anastasiadis2023-sf}
A.~Anastasiadis, L.~Paez, K.~Melo, E.~D. Tytell, A.~J. Ijspeert, and K.~Mulleners, ``\BIBforeignlanguage{en}{Identification of the trade-off between speed and efficiency in undulatory swimming using a bio-inspired robot},'' \emph{\BIBforeignlanguage{en}{Sci. Rep.}}, vol.~13, no.~1, p. 15032, Sep. 2023.

\bibitem{Zarrouk2016-ml}
D.~Zarrouk, M.~Mann, N.~Degani, T.~Yehuda, N.~Jarbi, and A.~Hess, ``\BIBforeignlanguage{en}{Single actuator wave-like robot ({SAW}): design, modeling, and experiments},'' \emph{\BIBforeignlanguage{en}{Bioinspir. Biomim.}}, vol.~11, no.~4, p. 046004, Jul. 2016.

\bibitem{Ta2023-jp}
T.~D. Ta and Y.~Kawahara, ``Printable bistable structures for programmable frictional skins of soft-bodied robots,'' in \emph{2023 IEEE/RSJ International Conference on Intelligent Robots and Systems (IROS)}.\hskip 1em plus 0.5em minus 0.4em\relax IEEE, Oct. 2023, pp. 8364--8370.

\end{thebibliography}

%%%%%%%%%%%%%%%%%%%%%%%%%%%%%%%%%%%%%%%%%%%%%%%%%%%%%%%%%%%%%%%%%%%%%%%%%%%%%%%%

\end{document}